\documentclass[runningheads]{llncs}

\usepackage{eccv}

\usepackage{eccvabbrv}

\usepackage{graphicx}
\usepackage{booktabs}
\usepackage{xcolor}
\usepackage{colortbl}
\usepackage{algorithm}
\usepackage{algorithmic}
\usepackage{bm}
\definecolor{pink}{rgb}{1, 0.8, 0.8}
\definecolor{orange}{rgb}{1, 0.9, 0.8}

\usepackage[accsupp]{axessibility}  % Improves PDF readability for those with disabilities.

\usepackage{hyperref}

\usepackage{orcidlink}

\begin{document}

\title{AVSplat: Dense-View Feed-Forward 3D Gaussian Splatting with Assist-View Preconditioning} 

\titlerunning{AVSplat: Dense-View Feed-Forward 3DGS}
\author{Muyu Xu\inst{1}\orcidlink{0009-0001-1547-2065} \and
Fangneng Zhan\inst{2}\orcidlink{0000-0003-1502-6847} \and
Yu Wei\inst{1}\orcidlink{2222--3333-4444-5555} \and \\
Hanspeter Pfister\inst{2}\orcidlink{0000-0002-3620-2582} \and
Shijian Lu\inst{1,*}\orcidlink{0000-0002-6766-2506}}

\authorrunning{M.~Xu et al.}

\institute{Nanyang Technological University, Singapore 639798, Singapore \\
\email{\{muyu001, ywei012\}@e.ntu.edu.sg} \\
\email{Shijian.Lu@ntu.edu.sg}\and
Harvard University, MA 02138, USA\\
\email{\{fnzhan, pfister\}@seas.harvard.edu}}

\maketitle
\begin{abstract}
Pose-free feed-forward 3D Gaussian Splatting enables novel view synthesis from uncalibrated multi-view images. Although more views should improve performance, existing methods often degrade with dense-view inputs because global aggregation spreads attention over many tokens, and naive voxel fusion averages many Gaussians into overly smooth representations. We present AVSplat, a framework that turns additional views into reliable signals for both aggregation and representation. Before global attention, each view performs a single lightweight interaction with a small set of Assist Views chosen for relevance and diversity, and the cached features provide a focused scene context that stabilizes correspondence. For representation, we use adaptive temperature-aware voxel fusion that sharpens attribution under high occupancy, guided by occupancy and point confidence. 
Crucially, AVSplat restores positive view scaling where performance remains stable or improves as more input views are added, instead of degrading in the dense-view regime. Ablations show that Assist View Preconditioning is primarily responsible for preventing dense-view degradation, while Occupancy-guided Voxel Fusion contributes most of the single-point image-quality gains.
\end{abstract}
    
\section{Introduction}
\label{sec:intro}
Recent feed-forward 3D Gaussian Splatting (3DGS)~\cite{smart2024splatt3r,ye2024no,zhang2025flare,kang2025selfsplat,jiang2025anysplat} enables pose-free novel view synthesis from uncalibrated multi-view images. A single forward pass predicts camera intrinsics and extrinsics together with a compact set of Gaussian primitives for differentiable rasterization.
These advances are based on geometric foundation models that infer structure directly from images. DUSt3R~\cite{wang2024dust3r} and MASt3R~\cite{leroy2024grounding} estimate dense point maps and pixel correspondences so downstream reconstruction can use point-level supervision without precomputed calibration. VGGT~\cite{wang2025vggt} estimates camera parameters, point maps, depth maps, and point tracks from one or many views and simplifies multi-view reasoning. Feed-forward 3DGS systems adopt such backbones and encode images with a geometry transformer before decoding features into Gaussian parameters and camera poses. This unifies geometric inference and radiance representation within a single pass.

Despite this progress, scaling from a few input views toward dense input views often yields a counterintuitive outcome in which reconstruction fidelity decreases as the number of input views increases~\cite{jiang2025anysplat, wang2025zpressor}. Two common design choices explain this behavior. The first lies in the global aggregation of the backbone~\cite{wang2025vggt} when the sequence length grows without known poses. In the absence of known poses, global aggregation relies on softmax attention to discover correspondences across many tokens. As the sequence length grows, the attention mass diffuses over many candidates, reducing the weight assigned to correct matches and weakening cross-view consistency~\cite{wang2025faster}.
The second lies in voxel fusion~\cite{jiang2025anysplat}. Per-pixel Gaussians are aggregated into a voxel grid to support efficient differentiable rasterization. When many Gaussians fall into the same voxel, a normalized average behaves as a low-pass filter. Strong contributors are diluted by many weak ones and high-frequency detail is lost. The issue becomes worse when the confidence of the contributor varies, but the fusion weights are not adaptive.
% Despite this progress, scaling from a few inputs to dense uncalibrated
% views can be counterintuitive. We identify two coupled failure modes.
% For a query token with a correct-match logit $\ell^\star$ and
% $N$ distractor logits $\{\ell_i\}_{i=1}^{N}$, the attention mass
% assigned to the correct match is
% \begin{equation}
% a^\star =
% \frac{\exp(\ell^\star)}
% {\exp(\ell^\star)+\sum_{i=1}^{N}\exp(\ell_i)}
% =
% \frac{1}
% {1+\sum_{i=1}^{N}\exp(\ell_i-\ell^\star)}.
% \end{equation}
% As more views are added, the distractor pool grows. Unless the
% correct-match margin increases accordingly, probability mass is diluted
% or transferred to visually similar false matches. At the representation
% stage, uniformly fusing slightly misregistered observations gives
% \begin{equation}
% \bar{\mathbf f}_n(\mathbf x)
% =
% \frac{1}{n}\sum_{i=1}^{n}
% \mathbf f(\mathbf x+\boldsymbol{\epsilon}_i)
% \xrightarrow[n\rightarrow\infty]{}
% \mathbb{E}_{\boldsymbol{\epsilon}}
% [\mathbf f(\mathbf x+\boldsymbol{\epsilon})],
% \end{equation}
% where $\boldsymbol{\epsilon}_i$ denotes a small registration error.
% The expectation corresponds to convolution with the misregistration
% distribution and therefore suppresses high-frequency content. These
% two effects motivate explicit control over both correspondence search
% and voxel attribution.

We propose AVSplat to improve global correspondence reasoning and voxel representation in the dense-view setting. We first propose an Assist View Preconditioning module. This module is designed to address a dense-view failure mode of prior pose-free pipelines, where the global attention becomes increasingly diffuse and cross-view matching becomes unstable as the number of input views grows. In this module, each current view performs a single lightweight interaction with a small Assist View set before any global attention. The Assist Views are both relevant to the current view and complementary to each other. Selection first finds strong neighbors by embedding similarity, then promotes diversity with a coverage-aware criterion, so that a small budget captures complementary observations and improves scene coverage. The features of Assist Views are cached for efficiency. The current view attends to these features and obtains conditioned tokens that carry the scene context before global aggregation. This focuses the correspondence search on plausible matches and reduces attention dispersion. A gated residual controls conditioning strength early in training and stabilizes optimization.

% AVSplat also introduces Occupancy-guided Voxel Fusion which includes a temperature-aware algorithm. The temperature depends on voxel occupancy and local consistency, so it reflects both the number of contributors and their reliability, and it will affect the sampling distribution. High occupancy lowers the temperature and sharpens the sampling distribution so that reliable contributors dominate the fused attributes. Low occupancy raises the temperature and keeps the distribution inclusive, so information is preserved without brittle winner-take-all behavior. The schedule is bound to keep the gradients stable. Multi-view agreement and photometric residuals modulate the temperature and limit the influence of outliers. This preserves a high-frequency structure where occupancy is high.
AVSplat also introduces Occupancy-guided Voxel Fusion, which adapts the fusion temperature to voxel occupancy and local consistency. High occupancy produces a sharper weight distribution, allowing reliable contributors to dominate the fused attributes. Under low occupancy, a smoother distribution retains information from multiple candidates and avoids brittle winner-take-all fusion. The temperature is constrained to a fixed range to stabilize optimization, while multi-view agreement and photometric residuals reduce the influence of outliers. This design helps preserve high-frequency structure in crowded voxels.

By converting additional views into a useful signal during both aggregation and representation, AVSplat preserves detail under high voxel occupancy and keeps correspondence search focused at long sequence length. In pose-free settings, it consistently improves overall performance with more input views under comparable compute and memory. Ablations confirm that Assist View Preconditioning and Occupancy-guided Voxel Fusion are both necessary for consistent gains. The analysis explains why prior feed-forward pipelines degrade when inputs become dense and shows that modeling attention sparsity and voxel attribution together leads to scalable and high-fidelity 3D Gaussian Splatting.

\noindent Our main contributions are:
\begin{itemize}
    \item We present AVSplat, a pose-free feed-forward framework that maintains positive view scaling in dense multi-view settings through improvements in global aggregation and voxel representation.
    \item We introduce Assist View Preconditioning, a lightweight per-view conditioning step that restores positive view scaling by preventing attention dilution and stabilizing cross-view matching at long sequence lengths.
    \item We propose Occupancy-guided Voxel Fusion, where the sampling distribution varies with the voxel occupancy and local consistency so that high-frequency detail is preserved.
    \item We conduct extensive experiments in pose-free settings, showing that
    AVSplat maintains or improves performance as the number of input views
    increases.
\end{itemize}
\section{Related Work}
\label{sec:related_work}

\subsection{Geometry foundation models}
Recent geometry foundation models predict dense point maps and correspondences directly from few views. DUSt3R~\cite{wang2024dust3r} recovers accurate point maps without calibration, providing strong priors for reconstruction and tracking. MASt3-R~\cite{leroy2024grounding} grounds the matching in 3D, producing robust, coarse-to-fine correspondences under wide baselines. MUSt3R~\cite{cabon2025must3r} extends to multi-view sequences and enables high-frame-rate inference for offline or online reconstruction. 
% VGGT~\cite{wang2025vggt} jointly infers cameras, depth, point maps, and tracks in one pass via alternating frame-wise and global aggregation, improving feed-forward reconstruction. However, global attention incurs quadratic cost and exhibits sparse patterns. Block-sparse alternatives preserve accuracy while accelerating long-sequence inference.
VGGT~\cite{wang2025vggt} advances the foundation model line by inferring camera parameters, depth, point maps, and point tracks in a single pass from one view to many views. The model alternates frame-wise and global aggregation and therefore reasons jointly over all frames. The approach improves the accuracy of feed-forward reconstruction and reduces reliance on multi-stage optimization. Analyses of global attention in VGGT and in related transformers reveal a dominant runtime cost that grows quadratically with sequence length and a sparse pattern in which a small set of patch interactions carries most probability mass. Replacing dense global attention with block sparse kernels preserves accuracy while accelerating inference on long sequences.

\subsection{Feed-forward 3D Gaussian Splatting}
Feed-forward 3DGS predicts scene-specific Gaussians and renders novel views without per-scene optimization. PixelSplat~\cite{charatan2024pixelsplat} learns from image pairs with a 3D probability field for stable training and real-time rendering. MVSplat~\cite{chen2024mvsplat} uses plane-sweep cost volumes to extract clean Gaussians from sparse multi-view inputs. DepthSplat~\cite{xu2025depthsplat} couples splatting with monocular depth, yielding mutual gains. MuSASplat~\cite{xu2026musasplat} improves pose-free sparse-view 3DGS through lightweight multi-scale adaptation and efficient cross-view feature fusion, while ZPressor~\cite{wang2025zpressor} compresses multi-view inputs into a compact latent state, enabling feed-forward 3DGS models to scale to over 100 input views.

Pose-free and COLMAP-free pipelines operate on uncalibrated inputs. Splatt-3R~\cite{smart2024splatt3r} transfers MASt3R priors to stereo; NoPoSplat~\cite{ye2024no} anchors Gaussians in a canonical camera space for real-time inference; SelfSplat~\cite{kang2025selfsplat} removes external 3D priors via self-supervised depth and pose. FLARE~\cite{zhang2025flare} features a cascaded learning paradigm with camera pose serving as the critical bridge, achieving accurate camera pose estimation and high-quality 3D rendering. PCR-GS~\cite{wei2025pcr} tackles COLMAP-free 3DGS from a per-scene optimization perspective through pose co-regularization. In contrast, AVSplat follows the one-pass feed-forward setting and focuses on dense-view scaling.

Recent methods adopt geometry transformers: AnySplat~\cite{jiang2025anysplat} encodes uncalibrated collections with a VGGT-style aggregator and predicts Gaussians and poses with voxel-level grouping. Scaling to dense inputs exposes two issues—diluted global attention and high-occupancy voxel fusion that smooths high-frequency details—which we address by strengthening correspondence before global aggregation and refining fusion in crowded voxels. 

Complementary 3DGS advances improve Gaussian representation, detail pre-servation, and controllable appearance. SOGS~\cite{zhang2025sogs} introduces second-order anchors to improve anchor-based 3DGS with compact anchor features and reduced model size. FreGS~\cite{zhang2024fregs} uses progressive frequency regularization to guide Gaussian densification from low to high frequencies and alleviate over-reconstruction. StyleGaussian~\cite{liu2024stylegaussian} extends Gaussian splatting to instant 3D style transfer through efficient feature rendering and a KNN-based 3D CNN decoder. These works are orthogonal to AVSplat, which targets dense-view aggregation and voxel attribution in pose-free feed-forward reconstruction.

\section{Method}
This section first recalls the principles of 3D Gaussian Splatting~\cite{kerbl20233d} and the transformer-style multi-view geometry encoder (VGGT~\cite{wang2025vggt}) that we build upon. We then present the overall pipeline of our method. The pipeline extends the previous feed-forward Gaussian predictor~\cite{jiang2025anysplat} with two new components. The first component performs assist view preconditioning so that every input view receives compact scene context before global aggregation, with the primary goal of preventing dense-view degradation caused by attention dispersion. The second component performs occupancy-guided temperature fusion so that voxel-level attributes remain sharp under dense inputs.
\begin{figure*}[t]
    \centering
    \includegraphics[width=\textwidth]{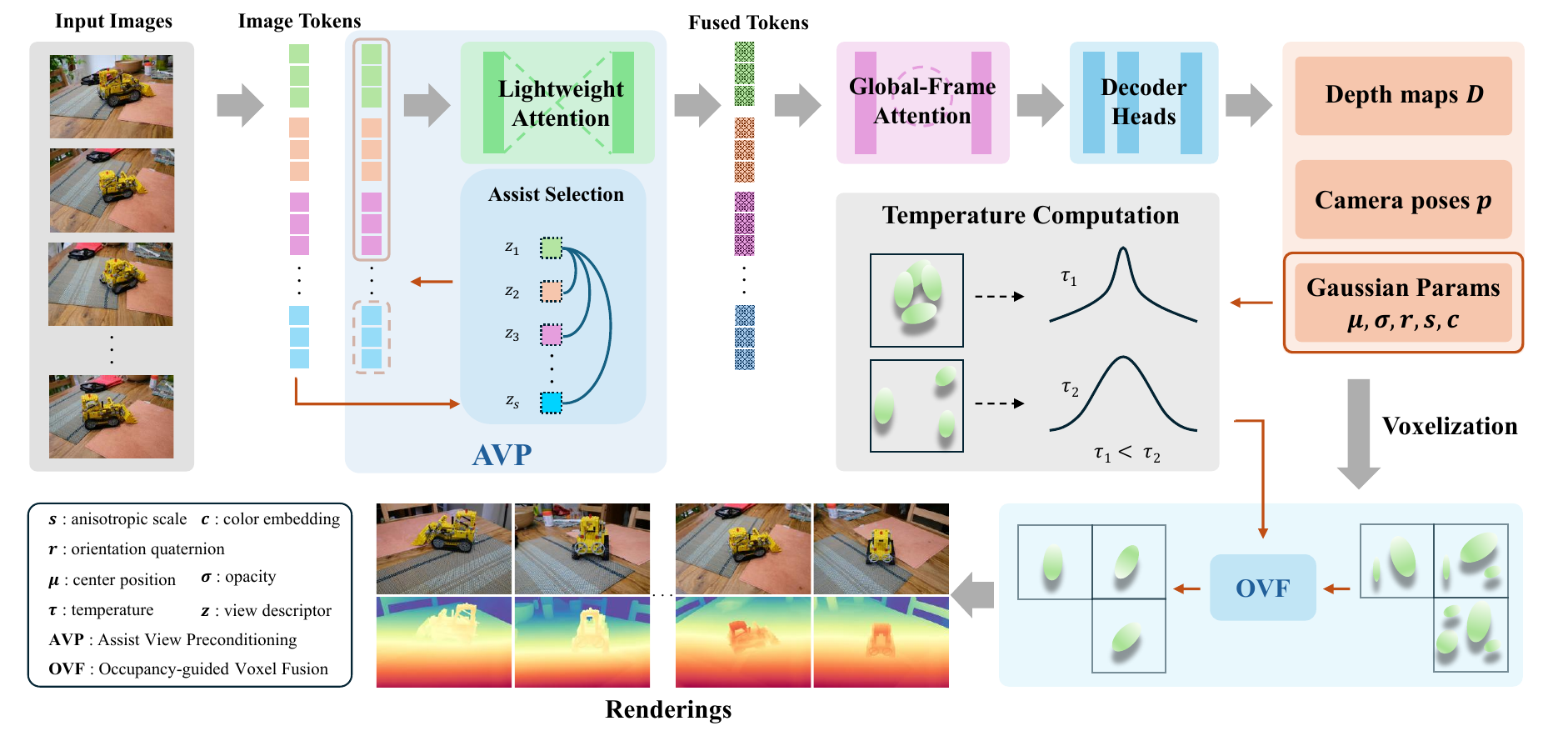}
    \caption{Overview of the AVSplat pipeline. The encoder converts each input image into patch embeddings (image tokens) and aggregates them into per-view tokens. Assist-View Preconditioning (AVP) computes a descriptor for each view and injects scene context before global aggregation. Decoder heads then predict camera parameters, depth, and Gaussian primitives. Occupancy-guided Voxel Fusion (OVF) forms voxel-level attributes and modulates them by voxel occupancy, followed by differentiable Gaussian rasterization to produce the final rendering.}
    \label{fig:pipeline}
\end{figure*}

\subsection{Preliminaries}

\subsubsection{3D Gaussian Splatting}
In 3D Gaussian Splatting (3DGS)~\cite{kerbl20233d}, a scene is represented by a finite set of anisotropic Gaussians
\begin{equation}
\mathcal{G}=\{G_i\}_{i=1}^{N}, \qquad
G_i=\bigl(\boldsymbol{\mu}_i,\boldsymbol{\Sigma}_i,\mathbf{c}_i,\alpha_i\bigr).
\end{equation}
Here $\boldsymbol{\mu}_i\in\mathbb{R}^3$ is the mean, $\boldsymbol{\Sigma}_i\in\mathbb{R}^{3\times 3}$ is a positive definite covariance, $\mathbf{c}_i\in\mathbb{R}^3$ is the radiance and $\alpha_i\in[0,1]$ is the opacity. For a calibrated camera with projection $\Pi$ and view direction $\mathbf{v}$, each Gaussian projects to the image plane as
\begin{equation}
\mathbf{u}_i=\Pi(\boldsymbol{\mu}_i)\qquad
\mathbf{S}_i=\mathbf{J}_i\,\boldsymbol{\Sigma}_i\,\mathbf{J}_i^{\top},
\end{equation}
where $\mathbf{J}_i$ is the Jacobian of $\Pi$ at $\boldsymbol{\mu}_i$. Let $\mathcal{S}(G_i,\mathbf{u})$ denote the induced 2D Gaussian kernel at pixel $\mathbf{u}$. Rendering proceeds in front-to-back order with color
\begin{equation}
\begin{aligned}
\mathbf{C}(\mathbf{u})
&=
\sum_{i=1}^{N}
T_i(\mathbf{u})\,\alpha_i\,\mathcal{S}(G_i,\mathbf{u})\,\mathbf{c}_i,
\end{aligned}
\end{equation}
and transmittance
\begin{equation}
T_i(\mathbf{u})=
\prod_{j<i}\bigl(1-\alpha_j\,\mathcal{S}(G_j,\mathbf{u})\bigr).
\end{equation}
Classical methods optimize $\mathcal{G}$ per scene from multi-view supervision. Feed-forward 3DGS removes per-scene optimization. A network takes images $\{I_s\}_{s=1}^{S}$ and directly predicts $\mathcal{G}$ together with camera intrinsics and extrinsics so that the rendering above supplies differentiable supervision in a single forward pass. The main challenges become multi-view feature aggregation and redundancy control in the predicted Gaussians.

\subsubsection{Transformer style multi-view geometry encoder}
In VGGT~\cite{wang2025vggt}, a geom-etry-aware encoder receives a sequence of images and produces tokens that carry appearance and 3D structure. For each view $s$ the encoder yields patch tokens
\begin{equation}
\mathbf{X}_s\in\mathbb{R}^{P\times d}, \quad s=1,\dots,S .
\end{equation}
Tokens are processed by alternating view-specific aggregation and global aggregation. The global stage operates on the concatenated set
\begin{equation}
\mathbf{X}=[\mathbf{X}_1,\dots,\mathbf{X}_S]\in\mathbb{R}^{(SP)\times d} .
\end{equation}
This allows information to flow across views when poses are unknown. Special tokens can be attached to support the prediction of camera parameters, depth, and point tracks. Global aggregation has quadratic complexity in the number of tokens, and its attention maps tend to become increasingly diffuse as sequence length grows. Our method addresses this weakness.

\subsection{AVSplat pipeline}
Given uncalibrated images, the encoder produces view tokens. A scene-level aggregation consumes all tokens and produces latents for camera estimation and Gaussian prediction. From these latents we decode intrinsics, extrinsics, Gaussian centers, scales, orientations, colors, and opacities. The resulting Gaussians are voxelized and rendered. An overview of our pipeline is shown in Figure~\ref{fig:pipeline}.

Our pipeline introduces two new designs. Before tokens enter global aggregation, we insert an assist view preconditioning module so that each view collects context from a compact set of assisting views. After Gaussian prediction, we replace uniform voxel fusion~\cite{jiang2025anysplat} with an occupancy-guided temperature fusion module that modulates fusion weights according to voxel occupancy and reliability. We refer to these modules as \emph{Assist View Preconditioning} and \emph{Occupancy-guided Voxel Fusion}.

\subsection{Assist View Preconditioning}
Global aggregation is expected to resolve correspondences across many views. When the number of views increases, tokens from unrelated views inflate the attention pool and the distribution becomes diffuse~\cite{wang2025faster}. Tokens from geometrically compatible views receive less probability mass and long-range constraints fail to strengthen the global match. Each view should first collect context from a compact and diverse set of assisting views and should enter global aggregation with this context.
\paragraph{Coverage driven assist selection}
For view $s$ with tokens $\mathbf{X}_s\in\mathbb{R}^{P\times d}$ we form a view descriptor by average pooling
\begin{equation}
\mathbf{z}_s=\frac{1}{P}\sum_{p=1}^{P}\mathbf{X}_s^{(p)}\in\mathbb{R}^{d}.
\end{equation}
We compute cosine similarity between views
\begin{equation}
m_{st}=\frac{\mathbf{z}_s^{\top}\mathbf{z}_t}{\lVert\mathbf{z}_s\rVert_2\,\lVert\mathbf{z}_t\rVert_2}.
\end{equation}
A candidate pool is formed by the $M$ nearest neighbors of $s$ in descriptor space
\begin{equation}
\mathcal{N}(s)=\operatorname{TopM}\,\bigl\{\,t\neq s\,:\,m_{st}\,\bigr\}.
\end{equation}

To measure how well a candidate view supports the tokens of $s$, we define a token-level support score. Let $\kappa(\mathbf{a},\mathbf{b})=\mathbf{a}^{\top}\mathbf{b}/(\lVert\mathbf{a}\rVert_2\,\lVert\mathbf{b}\rVert_2)$. Using a lightweight projection $\mathbf{W}_c\in\mathbb{R}^{d\times d}$, the support from view $t$ to token $p$ of $s$ is
\begin{equation}
s_{t,p}=\max_{q\in\{1,\dots,P\}}\,
\kappa\!\bigl(\mathbf{X}_s^{(p)}\mathbf{W}_c,\;\mathbf{X}_t^{(q)}\mathbf{W}_c\bigr).
\end{equation}
We then define a coverage objective over an assist set $\mathcal{A}\subset\mathcal{N}(s)$ of size $K$
\begin{equation}
\begin{aligned}
F(\mathcal{A})
&=\frac{1}{P}\sum_{p=1}^{P}\max_{t\in\mathcal{A}} s_{t,p}
\;+\;\beta\,\frac{1}{K}\sum_{t\in\mathcal{A}} q_t \\
&\quad-\;\lambda\,\frac{1}{K(K-1)}\!\!\sum_{\substack{t,u\in\mathcal{A}\\ t\neq u}} m_{tu},
\end{aligned}
\end{equation}
where $q_t$ is a view reliability term derived from token variance, the first term rewards token coverage, the second term prefers reliable helpers, and the third term discourages redundant views. Because the first term is monotone and submodular, greedy selection
provides a near-optimal solution at low computational cost. Starting from $\mathcal{A}_0=\varnothing$, we update
\begin{equation}
\mathcal{A}_{k+1}=\mathcal{A}_k\cup\bigl\{\arg\max_{t\in\mathcal{N}(s)\setminus\mathcal{A}_k}\,\Delta F(t\mid\mathcal{A}_k)\bigr\},
\end{equation}
until $\lvert\mathcal{A}\rvert=K$, where $\Delta F(t\mid\mathcal{A})=F(\mathcal{A}\cup\{t\})-F(\mathcal{A})$. In practice we evaluate $s_{t,p}$ on a stratified subset of tokens, cache per-view projections, and use random tie-breaking to improve generalization.

\paragraph{Lightweight conditioning}
Given the selected assist set $\mathcal{A}(s)$, we apply a single cross-view interaction to inject context. With linear maps absorbed into keys, queries, and values, the conditioned tokens are
\begin{equation}
\widetilde{\mathbf{X}}_s=\operatorname{softmax}\!\Bigl(\tfrac{1}{\sqrt{d}}\mathbf{Q}_s\mathbf{K}_s^{\top}\Bigr)\mathbf{V}_s,
\end{equation}
where $\mathbf{K}_s$ and $\mathbf{V}_s$ are formed by concatenating tokens from the assist set and $\mathbf{Q}_s$ is obtained from $\mathbf{X}_s$. We combine original and conditioned tokens through a gated residual
\begin{equation}
\mathbf{X}_s^{\text{out}}=\mathbf{X}_s+\gamma(g)\,\widetilde{\mathbf{X}}_s,
\end{equation}
where $g$ denotes the current training step and $\gamma(g)$ increases from near zero to one. The global aggregation then consumes $\{\mathbf{X}_s^{\text{out}}\}_{s=1}^{S}$.

This selection strategy favors views that cover different regions of the reference view and that agree with its content, while avoiding near duplicates. The single conditioning step moves the burden of discovery from the global module to the selection stage, so the subsequent aggregation focuses on long-range alignment rather than search. The resulting view-conditioned tokens are then fed into the same global aggregation and decoding heads as in the baseline.

\subsection{Occupancy-guided Voxel Fusion}
After obtaining all Gaussians, we project them onto a voxel grid so that Gaussians falling into the same or nearby voxels can be fused, reducing redundancy. Simple averaging within a voxel attenuates high-frequency appearance and smoo-ths fine-scale geometry, because numerous weak contributors dilute a few strong ones. The effect increases with occupancy and leads to over-smoothed reconstructions.

For voxel $v$ we define the candidate set
$
\mathcal{G}_v=\{G_i\}_{i\in\mathcal{I}_v},
$
and the occupancy
$
n_v=\lvert\mathcal{I}_v\rvert
$, where $\mathcal{I}_v$ is the index set of candidates that are assigned to voxel $v$ by the voxelization step.
Each candidate provides a feature vector $\mathbf{f}_i\in\mathbb{R}^{c}$, a 3D point $\mathbf{p}_i\in\mathbb{R}^{3}$ and a confidence $c_i\in\mathbb{R}_{\ge 0}$. We define a temperature
\begin{equation}
\begin{aligned}
\tau_v
=
\operatorname{clip}\!\Bigl(
\tau_{\min},\,
\tau_0\Bigl(\frac{n_v}{n_{\mathrm{ref}}}\Bigr)^{-\alpha}\,\psi_v,\,
\tau_{\max}
\Bigr) .
\end{aligned}
\end{equation}
with base temperature $\tau_0$, exponent $\alpha$ and bounds $\tau_{\min},\tau_{\max}$. The reliability factor $\psi_v$ depends on geometric consistency. We compute a variance
\begin{equation}
\sigma_v^2=\frac{1}{n_v}\sum_{i\in\mathcal{I}_v}\lVert\mathbf{p}_i-\bar{\mathbf{p}}_v\rVert_2^{2},
\end{equation}
with mean
$
\bar{\mathbf{p}}_v=\frac{1}{n_v}\sum_{i\in\mathcal{I}_v}\mathbf{p}_i ,
$
and set
\begin{equation}
\psi_v=\frac{1}{1+\lambda\,\sigma_v^{2}} .
\end{equation}
Here $\sigma_v^2$ measures the dispersion of 3D points within the voxel. Voxels with small variance are more geometrically consistent and therefore assigned a larger reliability factor $\psi_v$. We then form logits
$
\ell_i=\frac{c_i}{\tau_v}
$ for $i\in\mathcal{I}_v$
and compute softmax weights
\begin{equation}
w_i=\frac{\exp(\ell_i)}{\sum_{j\in\mathcal{I}_v}\exp(\ell_j)} .
\end{equation}
To reduce the influence of outliers, we sort candidates in descending order of $w_i$ and retain the shortest prefix whose cumulative weight reaches a target mass. The fused voxel feature and position are
\begin{equation}
\mathbf{f}_v=\sum_{i\in\mathcal{I}_v}w_i\,\mathbf{f}_i , \qquad
\mathbf{p}_v=\sum_{i\in\mathcal{I}_v}w_i\,\mathbf{p}_i .
\end{equation}

High occupancy reduces the temperature and sharpens the distribution so that candidates with high confidence or strong geometric agreement dominate. Low occupancy yields a higher temperature so that the fusion remains inclusive. The procedure preserves detail in well-observed regions and remains robust where support is limited. All steps are differentiable and trained jointly with the rest of the network.

Our method changes the order of information integration and the way voxel attribution is formed. Previous methods bring all tokens into a single pool and rely on a global module to discover relations. Assist view preconditioning makes each view context-aware before that stage. Standard fusion treats all candidates inside a voxel as equal competitors. Occupancy-guided Voxel Fusion makes the voxel sensitive to both how many candidates it receives and how consistent they are. These changes address the main causes of saturation in dense-view feed-forward 3DGS and restore the behavior that more views improve reconstruction quality.

\section{Experiments}
\label{sec:exp}
\subsection{Experimental setup}
\label{sec:exp_setup}
% ---------------------------- MAIN 2-COLUMN TABLE ----------------------------
\begin{table*}[!tb]
\centering
\caption{Quantitative results on Mip-NeRF 360~\cite{barron2022mip}, VR-NeRF~\cite{xu2023vr}, and DL3DV~\cite{ling2024dl3dv} under sparse and dense inputs with metrics PSNR$\uparrow$, SSIM$\uparrow$~\cite{wang2004image}, LPIPS$\downarrow$~\cite{zhang2018unreasonable}. The best result in each group is highlighted in red and the second best in orange. AVSplat attains rendering quality comparable to the strongest baselines and avoids the dense-view degradation observed in prior pose-free pipelines. Our performance remains stable or improves as more input views are added, instead of oscillating or degrading at high view counts.}
\resizebox{0.8\textwidth}{!}{
\begin{tabular}{l|ccc|ccc|ccc}
\toprule
% ==================== Sparse header ====================
& \multicolumn{9}{c}{Sparse inputs} \\
\cmidrule(lr){2-10}
& \multicolumn{3}{c}{3} & \multicolumn{3}{c}{6} & \multicolumn{3}{c}{16} \\
Method
& PSNR$\uparrow$ & SSIM$\uparrow$ & LPIPS$\downarrow$
& PSNR$\uparrow$ & SSIM$\uparrow$ & LPIPS$\downarrow$
& PSNR$\uparrow$ & SSIM$\uparrow$ & LPIPS$\downarrow$ \\
\midrule

\multicolumn{10}{c}{\textbf{Mip-NeRF 360}~\cite{barron2022mip}} \\
\midrule
NoPoSplat~\cite{ye2024no} &
16.36 & \cellcolor{orange}0.430 & 0.453 &
15.92 & 0.416 & 0.541 &
15.47 & 0.361 & 0.606 \\
Flare~\cite{zhang2025flare} &
13.52 & 0.350 & 0.601 &
15.35 & 0.407 & 0.573 &
13.21 & 0.348 & 0.695 \\
AnySplat~\cite{jiang2025anysplat} &
\cellcolor{orange}17.29 & 0.429 & \cellcolor{orange}0.433 &
\cellcolor{orange}17.51 & \cellcolor{orange}0.446 & \cellcolor{orange}0.412 &
\cellcolor{orange}17.81 & \cellcolor{pink}0.448 & \cellcolor{pink}0.382 \\
AVSplat &
\cellcolor{pink}17.33 & \cellcolor{pink}0.451 & \cellcolor{pink}0.354 &
\cellcolor{pink}17.99 & \cellcolor{pink}0.467 & \cellcolor{pink}0.352 &
\cellcolor{pink}18.13 & \cellcolor{orange}0.428 & \cellcolor{orange}0.390 \\
\midrule

\multicolumn{10}{c}{\textbf{VR-NeRF}~\cite{xu2023vr}} \\
\midrule
NoPoSplat~\cite{ye2024no} &
18.37 & 0.707 & 0.437 &
17.57 & 0.704 & 0.466 &
17.66 & \cellcolor{orange}0.720 & 0.472 \\
Flare~\cite{zhang2025flare} &
18.58 & 0.717 & 0.470 &
18.26 & 0.717 & 0.477 &
17.02 & 0.709 & 0.510 \\
AnySplat~\cite{jiang2025anysplat} &
\cellcolor{pink}20.41 & \cellcolor{orange}0.733 & \cellcolor{pink}0.389 &
\cellcolor{pink}21.07 & \cellcolor{orange}0.729 & \cellcolor{orange}0.436 &
\cellcolor{orange}21.12 & 0.714 & \cellcolor{pink}0.354 \\
AVSplat &
\cellcolor{orange}20.24 & \cellcolor{pink}0.764 & \cellcolor{orange}0.403 &
\cellcolor{orange}20.91 & \cellcolor{pink}0.736 & \cellcolor{pink}0.418 &
\cellcolor{pink}21.17 & \cellcolor{pink}0.724 & \cellcolor{orange}0.391 \\
\midrule

\multicolumn{10}{c}{\textbf{DL3DV}~\cite{ling2024dl3dv}} \\
\midrule
AnySplat~\cite{jiang2025anysplat} &
\cellcolor{pink}15.04 & \cellcolor{pink}0.387 & \cellcolor{orange}0.612 &
\cellcolor{pink}15.74 & \cellcolor{orange}0.420 & \cellcolor{pink}0.492 &
\cellcolor{orange}17.43 & \cellcolor{orange}0.542 & \cellcolor{orange}0.419 \\
AVSplat &
\cellcolor{orange}14.41 & \cellcolor{orange}0.341 & \cellcolor{pink}0.598 &
\cellcolor{orange}15.62 & \cellcolor{pink}0.424 & \cellcolor{pink}0.492 &
\cellcolor{pink}17.80 & \cellcolor{pink}0.565 & \cellcolor{pink}0.344 \\

% ==================== Dense header (same tabular) ====================
\midrule\midrule
& \multicolumn{9}{c}{Dense inputs} \\
\cmidrule(lr){2-10}
& \multicolumn{3}{c}{32} & \multicolumn{3}{c}{48} & \multicolumn{3}{c}{64} \\
% Method
% & PSNR$\uparrow$ & SSIM$\uparrow$ & LPIPS$\downarrow$
% & PSNR$\uparrow$ & SSIM$\uparrow$ & LPIPS$\downarrow$
% & PSNR$\uparrow$ & SSIM$\uparrow$ & LPIPS$\downarrow$ \\
\midrule

\multicolumn{10}{c}{\textbf{Mip-NeRF 360}~\cite{barron2022mip}} \\
\midrule
NoPoSplat~\cite{ye2024no} &
\multicolumn{3}{c|}{OOM} & \multicolumn{3}{c|}{OOM} & \multicolumn{3}{c}{OOM} \\
Flare~\cite{zhang2025flare} &
\multicolumn{3}{c|}{OOM} & \multicolumn{3}{c|}{OOM} & \multicolumn{3}{c}{OOM} \\
AnySplat~\cite{jiang2025anysplat} &
\cellcolor{orange}18.22 & \cellcolor{orange}0.445 & \cellcolor{orange}0.369 &
\cellcolor{pink}19.40 & \cellcolor{orange}0.476 & \cellcolor{pink}0.366 &
\cellcolor{orange}19.23 & \cellcolor{orange}0.500 & \cellcolor{pink}0.365 \\
AVSplat &
\cellcolor{pink}18.71 & \cellcolor{pink}0.483 & \cellcolor{pink}0.365 &
\cellcolor{orange}19.30 & \cellcolor{pink}0.482 & \cellcolor{orange}0.368 &
\cellcolor{pink}20.11 & \cellcolor{pink}0.528 & \cellcolor{orange}0.366 \\
\midrule

\multicolumn{10}{c}{\textbf{VR-NeRF}~\cite{xu2023vr}} \\
\midrule
NoPoSplat~\cite{ye2024no} &
\multicolumn{3}{c|}{OOM} & \multicolumn{3}{c|}{OOM} & \multicolumn{3}{c}{OOM} \\
Flare~\cite{zhang2025flare} &
\multicolumn{3}{c|}{OOM} & \multicolumn{3}{c|}{OOM} & \multicolumn{3}{c}{OOM} \\
AnySplat~\cite{jiang2025anysplat} &
\cellcolor{orange}21.11 & \cellcolor{orange}0.711 & \cellcolor{orange}0.380 &
\cellcolor{orange}21.60 & \cellcolor{pink}0.745 & \cellcolor{orange}0.338 &
\cellcolor{orange}21.26 & \cellcolor{orange}0.719 & \cellcolor{pink}0.350 \\
AVSplat &
\cellcolor{pink}21.42 & \cellcolor{pink}0.720 & \cellcolor{pink}0.365 &
\cellcolor{pink}21.62 & \cellcolor{orange}0.734 & \cellcolor{pink}0.317 &
\cellcolor{pink}21.7 & \cellcolor{pink}0.723 & \cellcolor{orange}0.363 \\
\midrule

\multicolumn{10}{c}{\textbf{DL3DV}~\cite{ling2024dl3dv}} \\
\midrule
AnySplat~\cite{jiang2025anysplat} &
\cellcolor{orange}21.05 & \cellcolor{orange}0.680 & \cellcolor{orange}0.284 &
\cellcolor{orange}21.04 & \cellcolor{orange}0.678 & \cellcolor{orange}0.281 &
\cellcolor{orange}21.14 & \cellcolor{orange}0.684 & \cellcolor{orange}0.282 \\
AVSplat &
\cellcolor{pink}21.54 & \cellcolor{pink}0.710 & \cellcolor{pink}0.266 &
\cellcolor{pink}21.60 & \cellcolor{pink}0.713 & \cellcolor{pink}0.265 &
\cellcolor{pink}21.67 & \cellcolor{pink}0.717 & \cellcolor{pink}0.265 \\

\bottomrule
\end{tabular}
}

\label{tab:main_results}
\end{table*}

% ---------------------------------------------------------------------------
% ---------------------------- 2-COLUMN FIGURE -------------------------------
% ---------------------------------------------------------------------------
\begin{figure*}[!tb]
    \centering
    \includegraphics[width=\textwidth]{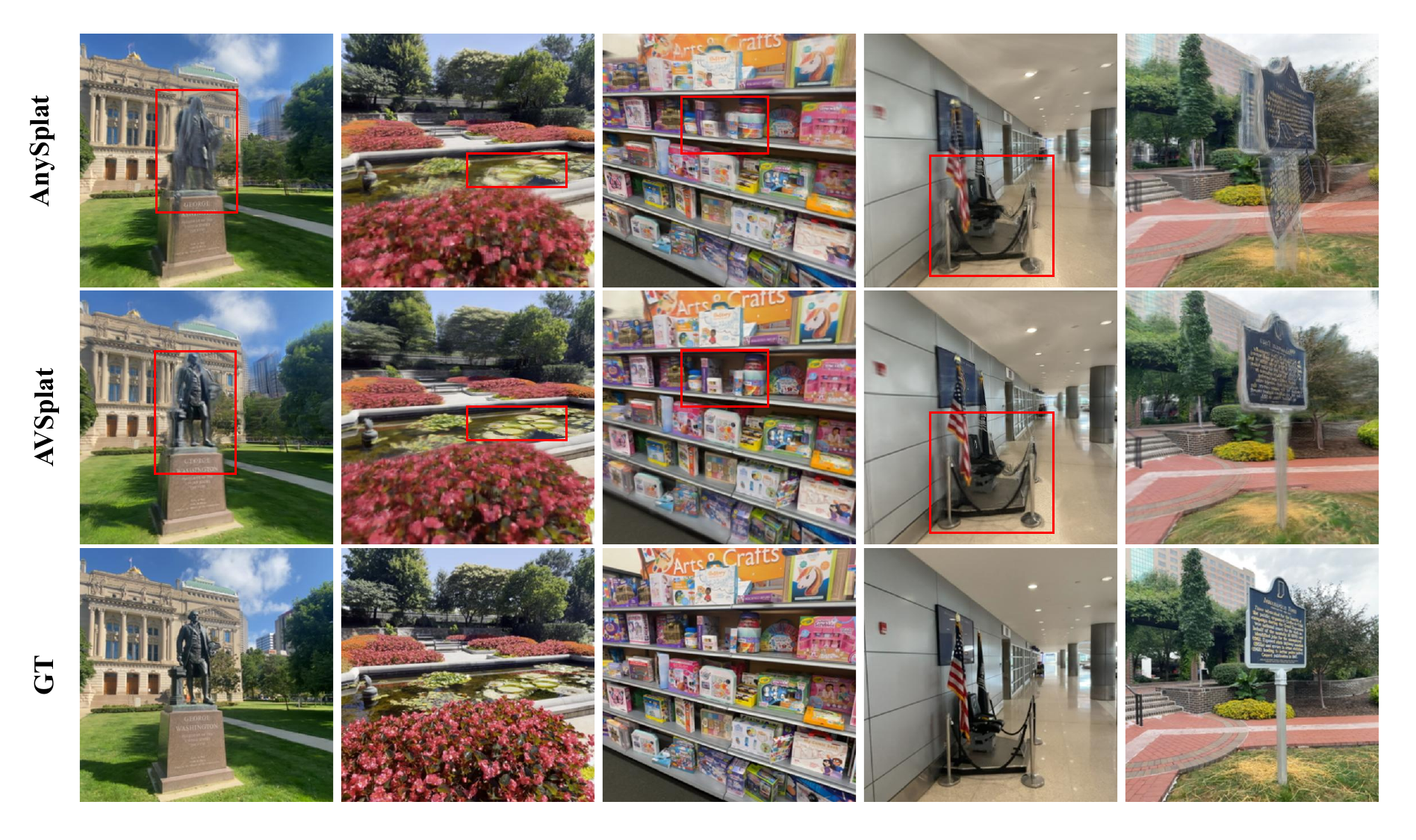}
    \caption{Qualitative comparison between AVSplat and AnySplat~\cite{jiang2025anysplat} under dense-view inputs. We focus on the dense-view setting, where the artifacts we target primarily occur, and therefore omit sparse-view results. Red boxes highlight that our method preserves more fine-grained surface details with fewer artifacts and less oversmoothing-induced blur, while the rightmost column shows more accurate global matching.}
    \label{fig:main_quali}
\end{figure*}
% ---------------------------- 2-COLUMN FIGURE -------------------------------
\begin{figure*}[!tp]
\centering
\includegraphics[width=0.95\textwidth]{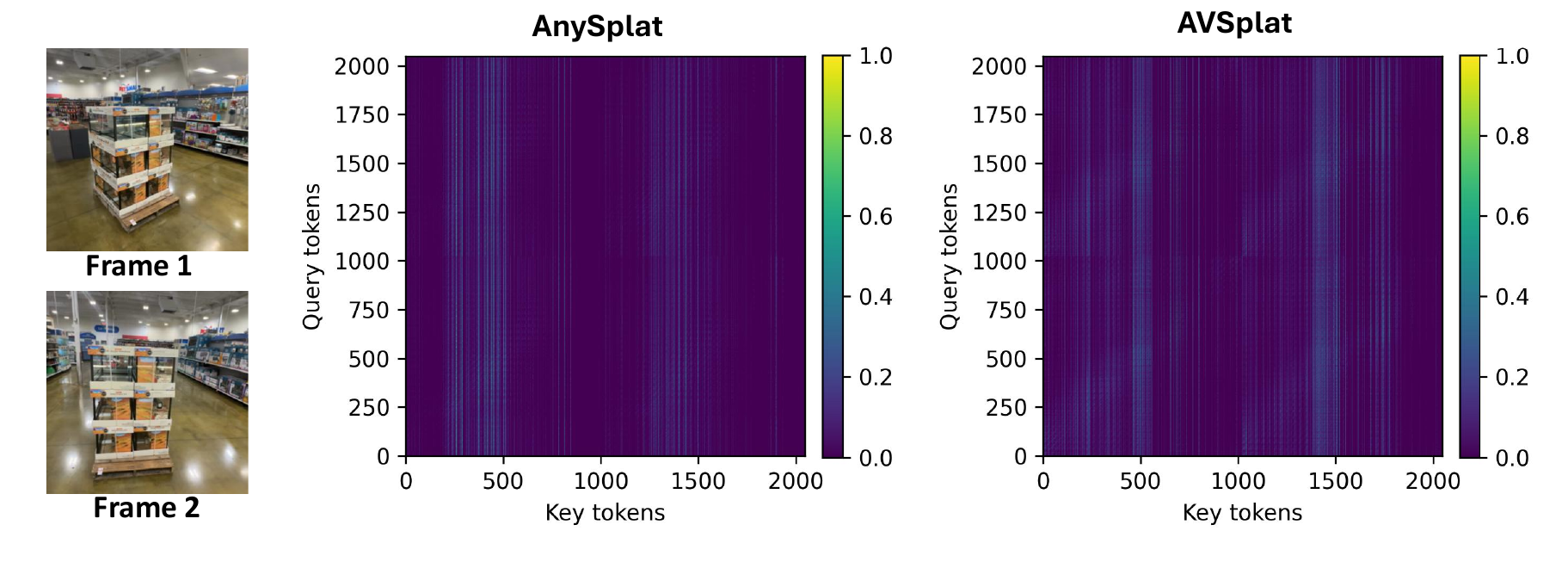}
\caption{Cross-view attention at layer 15 for a fixed frame pair under
16-view inputs. The maps are extracted from the full attention matrix
after removing self-frame blocks and special tokens, and are displayed
with identical normalization. AVSplat produces more concentrated
cross-view responses and the geometric consistency is quantified in
Tab.~\ref{tab:additional_diagnostics}.}
\label{fig:attn}
\end{figure*}

\begin{figure}[!tp]
\centering
\includegraphics[width=0.9\textwidth]{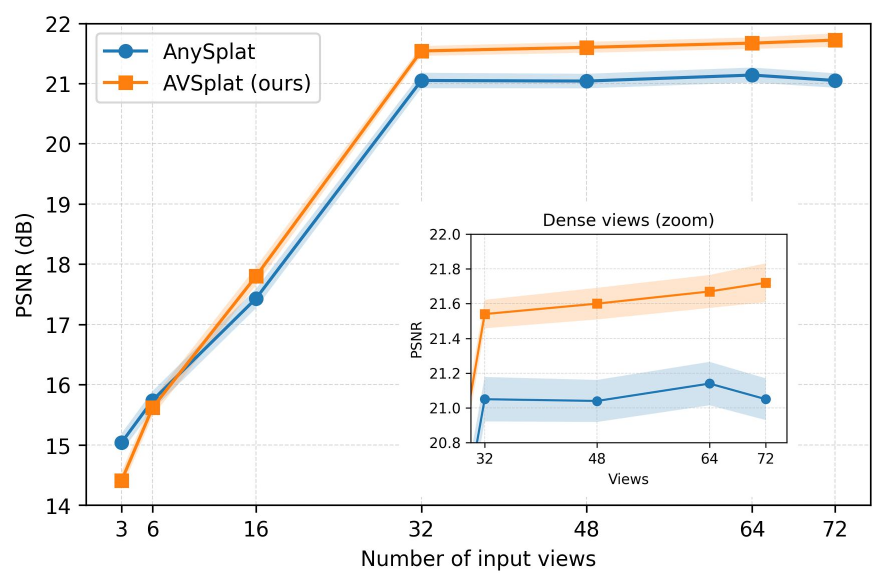}
% \caption{PSNR versus the number of input views. Shaded bands denote
% standard deviation across trials. AVSplat remains stable in the dense-view
% regime, whereas AnySplat oscillates at high view counts.}
\caption{Performance versus input count with 3, 6, 16, 32, 48, 64, and 72 input views. The vertical axis reports PSNR and the horizontal axis shows the number of input views. Shaded regions indicate the standard deviation across multiple trials. In the dense-view setting, AVSplat consistently outperforms AnySplat~\cite{jiang2025anysplat}. The inset in the lower right zooms into the dense-view region and shows that AnySplat begins to oscillate as the number of input views increases, whereas AVSplat maintains stable or improving performance without degradation.
}
\label{fig:curve}
\end{figure}

% ---------------------------------------------------------------------------
% ---------------------------- 2-COLUMN ABLATION TABLE -----------------------
\begin{table}[!htp]
\centering
\caption{Ablations on the DL3DV~\cite{ling2024dl3dv} in the 64 views setting. 
Occupancy-guided Voxel Fusion contributes most of the single-point image-quality improvement at 64 views, while Assist-View Preconditioning mainly targets the dense-view scaling failure mode by stabilizing global matching as the input count grows.}
\begin{tabular}{lccc}
\toprule
 Variant & PSNR $\uparrow$ & SSIM$\uparrow$ & LPIPS $\downarrow$ \\
\midrule
Baseline  & 21.14 & 0.684 & 0.282 \\
% \quad + Assist only          & 21.19 & 0.685 & 0.282 \\
\quad + Fusion only          & 21.57 & 0.692 & 0.272 \\
\quad + Assist + Fusion      & 21.67 & 0.717 & 0.265 \\
\bottomrule
\end{tabular}

\label{tab:ablate}
\end{table}

% ---------------------------------------------------------------------------
\begin{table}[t]
\centering
\caption{Additional dense-view diagnostics.
Left: cross-view attention analysis on DL3DV with 16 input views.
GeoMass@1patch measures the attention assigned to a one-patch-width
epipolar band. Right: PSNR (dB) on the overlap-heavy ACID evaluation.
Ground-truth poses are used only to compute the diagnostic metric and
are never provided to either model.}
\label{tab:additional_diagnostics}
\setlength{\tabcolsep}{4pt}
\begin{tabular}{lcc@{\qquad}lcc}
\toprule
\multicolumn{3}{c}{Attention diagnostics} &
\multicolumn{3}{c}{ACID overlap stress test} \\
\cmidrule(r){1-3}\cmidrule(l){4-6}
Method &
Entropy $\downarrow$ &
GeoMass $\uparrow$ &
Method &
48 views &
72 views \\
\midrule
AnySplat & 0.693 & 0.286 &
AnySplat & 23.12 & 22.45 \\
AVSplat  & \textbf{0.642} & \textbf{0.374} &
AVSplat  & \textbf{23.15} & \textbf{23.24} \\
\bottomrule
\end{tabular}
\end{table}

\paragraph{Datasets}
Our model is trained on the DL3DV dataset~\cite{ling2024dl3dv}, which provides over 10k training scenes. Novel view synthesis quality is evaluated on Mip-NeRF 360~\cite{barron2022mip}, VR-NeRF~\cite{xu2023vr}, and the DL3DV~\cite{ling2024dl3dv} test split. The DL3DV test split contains 140 scenes. For Mip-NeRF 360 and VR-NeRF we follow the settings used in prior work~\cite{jiang2025anysplat}. The sparse view setting uses 3, 6, and 16 input views per scene, and 2 target views per scene. The dense-view setting uses 32, 48, and 64 input views per scene, and 4, 6, and 8 target views, respectively. For VR-NeRF we use four indoor scenes that cover both compact and spacious layouts: apartment, kitchen, raf-furnishedroom, and workshop. For Mip-NeRF 360 we use four scenes with diverse viewpoints and lighting: bonsai, counter, kitchen, and room. These eight scenes span different room types, camera densities, and appearance patterns and are used for both sparse and dense-view evaluation. For each of the 140 DL3DV test scenes we first sample 75 candidate views, fix three random views as target views, and then randomly choose context views from the remaining views according to the desired input count.

\paragraph{Metrics and baselines.}
We report PSNR, SSIM~\cite{wang2004image}, and LPIPS~\cite{zhang2018unreasonable}. SSIM is computed with a Gaussian window in linear color space. LPIPS uses the VGG backbone with official weights. Scores are first averaged over target views within each scene and then averaged over scenes.
We compare with AnySplat~\cite{jiang2025anysplat} in the dense-view setting. The sparse view comparison includes NoPoSplat~\cite{ye2024no}, FLARE~\cite{zhang2025flare}, and AnySplat. Since evaluation on Mip-NeRF 360 and VR-NeRF is performed in a zero-shot setting, we retrain AnySplat on DL3DV using the public implementation and recommended configuration so that all methods share the same training data when comparing generalization performance.

\subsection{Implementation details}

\paragraph{Backbone and tokens}
The backbone is a transformer-style multi-view geometry encoder derived from VGGT~\cite{wang2025vggt}. Images are resized to a shorter side of 518 with preserved aspect ratio. The encoder uses a patch size of 14, an embedding dimension of 1024, and 24 layers with 16 attention heads. Each view yields a grid of patch tokens and a small set of special tokens. A global aggregation stage consumes tokens from all views.

\paragraph{Training schedule}
We train the model for 40,000 steps using AdamW with a learning rate of $1\times10^{-4}$ and a weight decay of $5\times10^{-2}$. We use a cosine learning-rate schedule with 2,000 warm-up steps. Total steps are 40000. Mixed precision and activation checkpointing are enabled. Gradient norm is clipped at 1. During training, the number of input images per scene is a random integer in $[4,24]$. Our model is trained with a batch size of 1 on a single NVIDIA A100 GPU with 80\,GB memory.

\paragraph{Assist View Preconditioning}
Each view selects 4 assisting views. We apply lightweight jittering with a probability of 0.05. The gating factor starts near 0 during a 2,000-step warm-up period and then increases linearly for 8000 steps until it reaches 1. We stop gradients through the assist branch for the first 1,500 steps. In the sparse-view setting, the number of available views is too small to form a meaningful assist set. Therefore, Assist-View Preconditioning is disabled in the sparse-view setting.

\paragraph{Occupancy-guided Voxel Fusion}
Voxel fusion uses a temperature that varies with occupancy and reliability. The base temperature is 0.7. The occupancy exponent is 0.5. The reference count is 8. The temperature is clipped to $[0.15, 2.0]$. The geometric-variance term is assigned a weight of 2.0. We use a top-mass threshold of 0.8 to truncate low-weight candidates.

\subsection{Results and analysis}
\paragraph{Main results across sparse and dense inputs}
% Table~\ref{tab:main_results} summarizes quantitative results on Mip-NeRF 360~\cite{barron2022mip}, VR-NeRF~\cite{xu2023vr}, and DL3DV~\cite{ling2024dl3dv} under sparse inputs $\{3,6,16\}$ and dense inputs $\{32,48,64\}$. Within each setting, the best and second-best scores are highlighted in red and orange, respectively. Across all datasets, AVSplat achieves performance that is comparable to or better than the strongest baselines, with particularly clear gains on DL3DV. On Mip-NeRF 360 and VR-NeRF, the margins are smaller, but still indicate that our model generalizes well across scenes and datasets. The advantage of AVSplat becomes more prominent in the dense-view setting, which is precisely where the artifacts we target are most severe. However, we also report sparse-view results to demonstrate that our model remains competitive when only a few input images are available. In this sparse setting, we disable the assist-view preconditioning module due to the limited number of input images, yet AVSplat still delivers strong performance. In the dense-view setting, AVSplat maintains a consistent improvement in rendering quality as the number of input views increases, whereas competing methods often exhibit oscillations or even degradation when many views are available. This behavior is consistent with our design of the assist-view preconditioning and the fusion mechanism, which are respectively intended to optimize global matching during token aggregation and to leverage high-occupancy voxels without oversmoothing fine structures.
Table~\ref{tab:main_results} reports results for sparse inputs with 3, 6, and 16 views and dense inputs with 32, 48, and 64 views. AVSplat remains competitive across all three datasets and shows its clearest gains on DL3DV. In the sparse setting, AVP is disabled because too few views are available to form a meaningful assist set. Under dense inputs, AVSplat exhibits more stable improvements as the view count increases, which is the setting targeted by AVP and occupancy-guided fusion.

Figure~\ref{fig:main_quali} provides a qualitative comparison between AVSplat and AnySplat~\cite{jiang2025anysplat} under dense-view inputs. Since the artifacts we target primarily arise in the dense-view setting, we only show qualitative comparisons for dense-view inputs here. Consistent with the quantitative trends, regions highlighted by red boxes show that AVSplat recovers sharper surface details with fewer artifacts and less oversmoothing-induced blur. The rightmost column further indicates that AVSplat produces more accurate global matching, leading to better large-scale structural consistency in the reconstructed scenes.

\paragraph{Dense-view scaling and overlap stress test.}
Under the fixed-target DL3DV protocol, AVSplat improves over AnySplat by 0.49, 0.56, and 0.53\,dB at 32, 48, and 64 input views, respectively. More importantly, its PSNR continues to increase from 21.54 to 21.67\,dB over this range, whereas the closest baseline largely saturates.

The difference becomes more pronounced for highly overlapping inputs. As shown in Table~\ref{tab:additional_diagnostics}, on ACID~\cite{liu2021infinite}, AnySplat decreases from 23.12\,dB at 48 views to 22.45\,dB at 72 views, whereas AVSplat increases from 23.15 to 23.24\,dB, producing a 0.79\,dB advantage at 72 views. The practical benefit is therefore not that every additional view produces a large single-step gain, but that redundant dense views no longer become harmful.

\paragraph{Attention analysis}
We visualize the activations of the global aggregation stage in Figure~\ref{fig:attn} to better understand how assist-view preconditioning affects cross-view matching. For a fixed pair of input frames, we extract the post-softmax attention matrix from layer 15 of the full global-aggregation sequence. We remove special tokens and self-frame blocks, and then retain the cross-frame block corresponding to the selected pair. Given these tokens, we compute the attention matrix as $\operatorname{softmax}(QK^\top)$, which yields a $2P \times 2P$ map where $P$ denotes the number of patch tokens per frame. 
The baseline exhibits diffuse attention mass when the number of views is large, indicating that global matching is poorly focused. In contrast, AVSplat produces attention patterns that are more sharply concentrated on geometrically plausible correspondences after preconditioning, demonstrating that our design effectively improves global matching.

\subsection{Ablation studies}
We ablate the Assist-View Preconditioning and Occupancy-guided Voxel Fusion modules in the 64-view setting on DL3DV~\cite{ling2024dl3dv}. Removing preconditioning reduces attention focus in the global aggregation stage and leads to lower PSNR and SSIM at long sequence lengths. Removing occupancy guidance causes excessive smoothing in high-occupancy regions, producing blur in the rendered images and suppressing fine surface details, which in turn degrades overall performance. As shown in Table~\ref{tab:ablate}, both components are effective. Occupancy-guided Voxel Fusion contributes most of the single-point image-quality improvement at 64 views, 
%while Assist-View Preconditioning plays a larger role in preventing performance degradation as the number of input views increases.
while Assist-View Preconditioning mainly targets the dense-view scaling failure mode by stabilizing global matching as the input count grows (see Figures~\ref{fig:attn} and~\ref{fig:curve}).

\paragraph{Performance vs. Number of input views}
Figure~\ref{fig:curve} further analyzes how performance scales with the number of input views. We plot PSNR for AVSplat and AnySplat using 3, 6, 16, 32, 48, 64, and 72 input views, with shaded bands indicating the standard deviation across multiple trials. Across the full range, both methods benefit substantially from moving from sparse to dense inputs. Beyond 32 views, however, their scaling behavior diverges. AnySplat begins to oscillate and slightly degrades at high view counts, whereas AVSplat continues to improve.
% The absolute gap between AVSplat and AnySplat is modest, but AVSplat consistently tracks a higher curve, especially in the dense-view setting.
The absolute gap at a fixed view count can be modest, but the scaling behavior differs consistently, which is the primary effect we target with Assist-View Preconditioning. 
% AVSplat tracks a higher and more stable curve, especially in the dense-view setting. In the dense-view setting (32 views and above), however, the behavior diverges: the AnySplat curve begins to oscillate and even slightly degrades at high view counts, whereas AVSplat continues to improve monotonically without degradation. 
This shows that AVSplat can consistently exploit additional views in dense settings, and complements the quantitative trends observed in Table~\ref{tab:main_results}.

\paragraph{Quantitative attention diagnostics.}
To quantify the effect of AVP beyond the qualitative heatmaps in Fig.~\ref{fig:attn}, we extract the post-softmax attention matrix from layer 15 of the full global-aggregation sequence. We exclude self-frame blocks and special tokens and average the results over four fixed view pairs. For cross-view attention $a_{qk}$, normalized entropy is the per-query entropy divided by the logarithm of the number of valid key patches. GeoMass@1patch is defined as $\operatorname{GeoMass}=\frac{1}{|\mathcal Q|}\sum_{q\in\mathcal Q}\sum_{k\in\mathcal E(q)} a_{qk},$ where $\mathcal E(q)$ contains key patches within a one-patch-width epipolar band computed from ground-truth evaluation poses.  
As shown in Table~\ref{tab:additional_diagnostics}, AVSplat reduces normalized entropy from 0.693 to 0.642 and increases GeoMass@1patch from 0.286 to 0.374, corresponding to an absolute gain of 8.8 points and a relative gain of 30.8\%. The mass of a random band is approximately 0.10, so the concentration on geometrically compatible regions increases from approximately $2.9\times$ to $3.7\times$ the random baseline.

\section{Conclusion}
This work addresses dense-view degradation in pose-free feed-forward 3D Gaussian Splatting. Assist-View Preconditioning conditions each view before global aggregation, while Occupancy-guided Voxel Fusion adapts voxel weights to occupancy and reliability. Experiments on Mip-NeRF 360,VR-NeRF, and DL3DV show more stable performance as the number of input views increases, and the attention analysis and ablations support the contributions of the two modules. Together, these changes improve dense-view pose-free reconstruction without per-scene optimization.
% This work addresses two factors that hinder scaling in feed-forward 3D Gaussian Splatting. When the number of views grows, global aggregation often diffuses attention, and naive voxel fusion attenuates high-frequency content in crowded regions. AVSplat tackles these issues with Assist-View Preconditioning and
% Occupancy-guided Voxel Fusion. The first module injects scene context before global aggregation so that correspondence search remains focused as sequence length increases. The second module adapts fusion temperature to voxel occupancy and reliability so that strong contributors dominate when evidence is abundant and information is preserved when support is weak.
% On MipNeRF-360~\cite{barron2022mip}, VR-NeRF~\cite{xu2023vr}, and DL3DV~\cite{ling2024dl3dv}, AVSplat restores the expected scaling trend, with additional input views generally maintaining or improving reconstruction quality in the dense-view setting. Attention maps show sharper cross-view matching after preconditioning, and ablations confirm that occupancy-aware fusion preserves detail at high density. These changes restructure information flow and voxel attribution and enable scalable pose-free reconstruction.
% Future work includes learned view selection with differentiable subset choice, sparse or deformable attention for very long sequences, uncertainty-aware rendering for reflective and transparent regions, and temporal extensions for dynamic scenes.

\section*{Acknowledgements}
This work was supported in part by the Ministry of Education, Singapore, under the Tier-2 project scheme (Project No.~MOE-T2EP20123-0003), and in part by the U.S. National Science Foundation under Grant No.~\mbox{CRCNS-2309041}.

% Please insert your acknowledgments here.
\clearpage
\bibliographystyle{splncs04}
\bibliography{main}

\clearpage
\setcounter{page}{1}

\phantomsection
\begin{center}
    {\LARGE\bfseries Supplementary Material\par}
    \vspace{0.5em}
\end{center}
\vspace{1em}

\setcounter{table}{0}  
\setcounter{figure}{0}

\section{Additional Algorithmic Details}

\subsection{Assist-View Preconditioning (Sec.~3.3)}

Before any global aggregation, we precondition the per-view patch tokens by letting each view attend to a small set of assist views. For a batch element, let
$\mathbf{X}_s \in \mathbb{R}^{P \times C}$ denote the patch-token grid of view $s \in \{1,\dots,S\}$, with $P$ spatial tokens and embedding dimension $C$. We first compute a descriptor
$\mathbf{d}_s = \mathrm{norm}\big(\frac{1}{P} \sum_{p=1}^{P} \mathbf{X}_s[p]\big)$
for each view and a subsampled token set $\hat{\mathbf{X}}_s \in \mathbb{R}^{\hat{P} \times C}$ for coverage, obtained by regularly sampling $P_{\text{sub}}$ tokens and $\ell_2$ normalizing them.

For each target view $s$, we construct a candidate pool $\mathcal{N}(s)$ using $M$ nearest neighbors in descriptor space (cosine similarity). Among these candidates, we greedily pick $K$ assist views by maximizing a score that combines:
(i) similarity of the candidate descriptor to the target descriptor,
(ii) diversity with respect to already selected assist views in descriptor space,
(iii) token-level coverage gain on $\hat{\mathbf{X}}_s$, and
(iv) a reliability score derived from within-view token variance.
The greedy procedure runs on detached tokens and descriptors, so the selection is treated as a non-differentiable side computation.

Given the selected assist set $\mathcal{A}(s)$, we perform one multi-head attention (MHA) per view with queries from the current view and keys/values from the assist views. The resulting context is added back as a gated residual:
\[
\tilde{\mathbf{X}}_s = \mathbf{X}_s + \alpha(g)\,\mathrm{stopgrad}\big(\mathbf{C}_s\big)
\quad\text{for } g < g_{\text{stop}},
\]
and $\tilde{\mathbf{X}}_s = \mathbf{X}_s + \alpha(g)\mathbf{C}_s$ afterwards, where $g$ is the global training step, $\alpha(g)$ is a piecewise-linear gate that ramps from $0$ to $1$, and $\mathbf{C}_s$ is the MHA output. In our implementation we use $K=4$, a warm-up of $2000$ steps, a ramp of $8000$ steps, and stop gradients from the assist-view branch for the first $1500$ steps.

\begin{algorithm}[t]
\caption{Assist-View Preconditioning}
\label{alg:assist_view}
\footnotesize
\begin{algorithmic}[1]
\REQUIRE Per-view tokens $\{X_s \in \mathbb{R}^{P\times C}\}_{s=1}^S$, \\
         number of assist views $K$, candidate pool size $M$, \\
         subsample size $P_{\text{sub}}$, diversity weight $\lambda_{\text{div}}$, \\
         reliability weight $\beta$, gate schedule $\alpha(g)$, stop-grad step $g_{\mathrm{stop}}$
\ENSURE Conditioned tokens $\{\tilde{X}_s\}_{s=1}^S$
\vspace{0.2em}

\STATE \textbf{View descriptors and reliability}
\FOR{$s = 1$ {\bf to} $S$}
  \STATE Compute descriptor $d_s = \mathrm{norm}\!\left(\frac{1}{P}\sum_{p} X_s[p]\right)$.
  \STATE Subsample $P_{\text{sub}}$ tokens from $X_s$ and compute a variance-based reliability $q_s \in [0,1]$.
\ENDFOR
\STATE Stack $D = [d_1,\dots,d_S]^\top$; similarity matrix $\Sigma = \mathrm{clip}(DD^\top,-1,1)$; distance matrix $\Delta = \max(1-\Sigma,0)$.

\vspace{0.2em}
\STATE \textbf{Assist-view selection}
\FOR{$s = 1$ {\bf to} $S$}
  \STATE Candidate pool $N(s)$: indices of the top-$M$ neighbors of $s$ in $\Sigma$ (excluding $s$).
  \STATE Initialize selected assist set $A(s)$ with the most similar neighbor in $N(s)$ and initialize coverage scores on the subsampled tokens of view $s$.
  \WHILE{$|A(s)| < K$ \AND $|A(s)| < |N(s)|$}
    \STATE For each $t \in N(s)\setminus A(s)$, compute
    \STATE \hspace{1em} similarity term from $\Sigma[s,t]$,
           diversity term from distances $\Delta[t,u]$ for $u \in A(s)$,
           coverage gain on the subsampled tokens of $s$ when adding $t$,
           and reliability term $q_t$.
    \STATE Combine them into
           $\text{score}(t) = 0.7\,\text{sim}(t) + \lambda_{\text{div}}\,\text{div}(t) + \text{cov}(t) + \beta\,q_t$.
    \STATE Select $t^\star = \arg\max_{t} \text{score}(t)$; update $A(s) \leftarrow A(s)\cup\{t^\star\}$ and coverage scores.
  \ENDWHILE
  \STATE Optionally apply a small random jitter by replacing the last assist in $A(s)$ with another candidate.
\ENDFOR

\vspace{0.2em}
\STATE \textbf{Cross-attention and gated residual}
\STATE Set $\alpha \leftarrow \alpha(g)$ using the warm-up and ramp schedule.
\FOR{$s = 1$ {\bf to} $S$}
  \STATE Form query $Q = X_s$ and keys/values by concatenating $\{X_t\}_{t \in A(s)}$ along the token dimension.
  \STATE Compute assist context $C_s = \mathrm{MHA}(Q,K,V)$.
  \IF{$g < g_{\mathrm{stop}}$}
    \STATE Detach gradients: $C_s \leftarrow \mathrm{stopgrad}(C_s)$.
  \ENDIF
  \STATE $\tilde{X}_s \leftarrow X_s + \alpha\,C_s$.
\ENDFOR
\STATE \textbf{return} $\{\tilde{X}_s\}_{s=1}^S$
\end{algorithmic}
\end{algorithm}
\begin{algorithm}[t]
\caption{Occupancy-Guided Voxel Fusion}
\label{alg:voxel_fusion}
\begin{algorithmic}[1]
\REQUIRE Points $P$, features $F$, optional confidence map $C$, voxel size $\Delta$, \\
         base temperature $\tau_0$, exponent $\alpha$, bounds $(\tau_{\min},\tau_{\max})$, \\
         reference count $n_{\text{ref}}$, reliability weights $(\beta_{\text{conf}},\beta_{\text{geo}})$, \\
         reliability factor $k_{\text{rel}}$, optional trim mass $m_{\text{th}}$
\ENSURE Fused voxel centers $\{p_v\}$ and features $\{f_v\}$
\STATE Flatten $P,F$ (and $C$ if available) to samples $(p_i,f_i,c_i)$, $i=1,\dots,N$.
\STATE Map $c_i$ to logits $\ell_i$ and probabilities $q_i \in (0,1)$; if $C$ is absent set $\ell_i = 0$, $q_i = 0.5$.
\STATE Compute voxel indices $k_i = \mathrm{round}(p_i / \Delta)$ and group indices into voxel sets $I_v = \{ i \mid k_i = k_v \}$.
\FOR{each voxel $v$}
  \STATE $n_v \leftarrow |I_v|$, \quad
         $\mu_v \leftarrow \tfrac{1}{n_v} \sum_{i \in I_v} p_i$, \quad
         $\sigma_v^2 \leftarrow \tfrac{1}{n_v} \sum_{i \in I_v} \|p_i - \mu_v\|_2^2$.
  \STATE $g_v \leftarrow \tfrac{1}{1 + \sigma_v^2 / (\Delta^2 + \varepsilon)}$, \quad
         $\bar c_v \leftarrow \tfrac{1}{n_v} \sum_{i \in I_v} q_i$.
  \STATE $r_v \leftarrow \mathrm{clip}\!\Big(
         \tfrac{\beta_{\text{conf}}\bar c_v + \beta_{\text{geo}} g_v}
              {\beta_{\text{conf}} + \beta_{\text{geo}} + \varepsilon},
         0, 1 \Big)$.
  \STATE $\tilde n_v \leftarrow \max(n_v / n_{\text{ref}}, \varepsilon)$, \quad
         $\tau_v^{\text{occ}} \leftarrow \tau_0 \tilde n_v^{-\alpha}$, \quad
         $u_v \leftarrow \tfrac{1}{1 + k_{\text{rel}} (1 - r_v)}$, \quad
         $\tau_v \leftarrow \mathrm{clip}(\tau_v^{\text{occ}} u_v, \tau_{\min}, \tau_{\max})$.
  \STATE For $i \in I_v$ set $z_i = \ell_i / \tau_v$ and obtain weights $w_i$ from softmax or sparsemax over $\{z_i\}_{i \in I_v}$.
  \IF{trimming is enabled}
    \STATE Keep the smallest prefix of $\{w_i\}_{i \in I_v}$ whose cumulative sum $\ge m_{\text{th}}$ (at least one element) and renormalize.
  \ENDIF
  \STATE $p_v \leftarrow \sum_{i \in I_v} w_i p_i$, \quad
         $f_v \leftarrow \sum_{i \in I_v} w_i f_i$.
\ENDFOR
\STATE \textbf{return} $\{p_v\}_v, \{f_v\}_v$.
\end{algorithmic}
\end{algorithm}

\subsection{Occupancy-Guided Voxel Fusion (Sec.~3.4)}

We fuse per-pixel 3D points and features into a sparse set of voxels using an occupancy- and reliability-aware temperature. Let
$\mathbf{p}_i \in \mathbb{R}^3$ and $\mathbf{f}_i \in \mathbb{R}^C$
denote the 3D position and feature of sample $i$ (coming from all views and pixels), and let $c_i$ be an optional confidence. We discretize the positions into voxels of size $\Delta$ via
$\mathbf{k}_i = \mathrm{round}(\mathbf{p}_i / \Delta)$
and group samples by voxel index. For each voxel $v$, we compute:

\begin{itemize}
  \item occupancy $n_v$,
  \item mean position $\boldsymbol{\mu}_v$ and normalized variance $\sigma^2_v$,
  \item geometric consistency $g_v = 1 / (1 + \sigma^2_v)$,
  \item mean confidence $\bar{c}_v$ (after mapping to probability if needed),
  \item reliability $r_v \in [0,1]$ as a weighted combination of $\bar{c}_v$ and $g_v$.
\end{itemize}

The voxel temperature is then
\[
\tau_v = \mathrm{clip}\!\left(
  \tau_{\min},\;
  \tau_0 \Big(\frac{n_v}{n_{\text{ref}}}\Big)^{-\alpha} \cdot
  \frac{1}{1 + k_{\text{rel}} (1 - r_v)},\;
  \tau_{\max}
\right),
\]
so dense, reliable voxels use a smaller temperature and produce sharper weights, while sparse or unreliable voxels use a larger temperature to avoid overconfident fusion. For each voxel we apply a per-voxel softmax (or sparsemax) with logits scaled by $\tau_v$, optionally trim a prefix of weights until a cumulative mass threshold (e.g., $0.8$), and renormalize. The fused voxel position and feature are the weighted averages under these final weights.

\section{More qualitative results}
We provide more qualitative results in the format of videos. Please refer to the video in the supplementary materials.

% \section{Extra experiment}
% \begin{table}[!htp]
% \centering
% \setlength{\tabcolsep}{6pt}
% \caption{More ablations on the dense setting at 64 views. $K$ is the number of assisting views. $\alpha$ is the occupancy exponent. $\tau$ is the temperature range. $M$ is the top-mass threshold. Each cell reports a placeholder.}
% \label{tab:ablate}
% \scriptsize
% \begin{tabular}{lcccccc}
% \toprule
% Variant & $K$ & $\alpha$ & $\tau$ & $M$ & PSNR $\uparrow$ & LPIPS $\downarrow$ \\
% \midrule

% AVSplat      & 4  & 0.5 & $[0.15,2.0]$ & 0.8 & 21.67 & 0.265 \\
% \midrule
% Assist count sweep           & 2  & 0.5 & $[0.15,2.0]$ & 0.8 & -- & -- \\
%                              & 4  & 0.5 & $[0.15,2.0]$ & 0.8 & -- & -- \\
%                              & 6  & 0.5 & $[0.15,2.0]$ & 0.8 & -- & -- \\
% \midrule
% Fusion exponent sweep        & 4  & 0.3 & $[0.15,2.0]$ & 0.8 & -- & -- \\
%                              & 4  & 0.5 & $[0.15,2.0]$ & 0.8 & -- & -- \\
%                              & 4  & 0.8 & $[0.15,2.0]$ & 0.8 & -- & -- \\
% \midrule
% Top-mass sweep               & 4  & 0.5 & $[0.15,2.0]$ & 0.6 & -- & -- \\
%                              & 4  & 0.5 & $[0.15,2.0]$ & 0.8 & -- & -- \\
%                              & 4  & 0.5 & $[0.15,2.0]$ & 0.9 & -- & -- \\
% \bottomrule
% \end{tabular}
% \end{table}

\end{document}